\documentclass[letterpaper, 10 pt, conference]{ieeeconf}  % Comment this line out if you need a4paper
\usepackage{graphicx}
\usepackage{multirow}
\usepackage{amsmath,amssymb,amsbsy}
\usepackage{graphicx}
\usepackage{caption}
\usepackage{svg}
\usepackage{hyperref}

\hypersetup{
    colorlinks=true,
    linkcolor=blue,
    filecolor=magenta,      
    urlcolor=blue,
    pdftitle={SKoPe3D},
    pdfpagemode=FullScreen,
}

\IEEEoverridecommandlockouts                              % This command is only needed if 
\title{\LARGE \bf
SKoPe3D: A Synthetic Dataset for Vehicle Keypoint Perception in 3D from Traffic Monitoring Cameras
}

\author{Himanshu Pahadia$^{1*}$, Duo Lu$^{2*}$, 
Bharatesh Chakravarthi$^{1}$, Yezhou Yang$^{1}$% <-this % stops a space
\\
Project Dataset: \url{https://duolu.github.io/skope3d.html}
\thanks{This research is sponsored by the Institute of Automated Mobility (Arizona) and the US National Science Foundation Pathways to Enable Open-Source Ecosystems grant (\#2303748).}% <-this % stops a space
\thanks{$^{1}$School of Computing and Augmented Intelligence, Arizona State University, Tempe, AZ, \{hpahadia, bshettah, yz.yang\}@asu.edu}%
\thanks{$^{2}$Rider University, Lawrenceville, NJ, dlu@rider.edu}%
\thanks{$^{*}$Equal contribution.}%
}

\begin{document}

% \thanks{$^{1}$Albert Author is with Faculty of Electrical Engineering, Mathematics and Computer Science,
%         University of Twente, 7500 AE Enschede, The Netherlands
%         {\tt\small albert.author@papercept.net}}%

\maketitle
\thispagestyle{empty}
\pagestyle{empty}

%%%%%%%%%%%%%%%%%%%%%%%%%%%%%%%%%%%%%%%%%%%%%%%%%%%%%%%%%%%%%%%%%%%%%%%%%%%%%%%%
\begin{abstract}

Intelligent transportation systems (ITS) have revolutionized modern road infrastructure, providing essential functionalities such as traffic monitoring, road safety assessment, congestion reduction, and law enforcement. Effective vehicle detection and accurate vehicle pose estimation are crucial for ITS, particularly using monocular cameras installed on the road infrastructure. One fundamental challenge in vision-based vehicle monitoring is keypoint detection, which involves identifying and localizing specific points on vehicles (such as headlights, wheels, taillights, etc.). However, this task is complicated by vehicle model and shape variations, occlusion, weather, and lighting conditions. Furthermore, existing traffic perception datasets for keypoint detection predominantly focus on frontal views from ego vehicle-mounted sensors, limiting their usability in traffic monitoring. To address these issues, we propose SKoPe3D, a unique synthetic vehicle keypoint dataset generated using the CARLA simulator from a roadside perspective. This comprehensive dataset includes generated images with bounding boxes, tracking IDs, and 33 keypoints for each vehicle. Spanning over 25k images across 28 scenes, SKoPe3D contains over 150k vehicle instances and 4.9 million keypoints. To demonstrate its utility, we trained a keypoint R-CNN model on our dataset as a baseline and conducted a thorough evaluation. Our experiments highlight the dataset's applicability and the potential for knowledge transfer between synthetic and real-world data. By leveraging the SKoPe3D dataset, researchers and practitioners can overcome the limitations of existing datasets, enabling advancements in vehicle keypoint detection for ITS. 

\end{abstract}

%%%%%%%%%%%%%%%%%%%%%%%%%%%%%%%%%%%%%%%%%%%%%%%%%%%%%%%%%%%%%%%%%%%%%%%%%%%%%%%%
\section{INTRODUCTION}
%%%%%%%%%%%%%%%%%%%%%%%%%%%%%%%%%%%%%%%%%%%%%%%%%%%%%%%%%%%%%%%%%%%%%%%%%%%%%%%%

As the number of vehicles on our roadways continues to surge, a host of road traffic-related challenges have arisen, including traffic congestion, accidents, infrastructure strain, transportation inefficiency, and environmental pollution. To tackle these pressing issues, researchers are actively developing intelligent transportation systems (ITS) that integrate sophisticated traffic monitoring capabilities using cameras. By employing advanced computer vision methods such as neural networks, these systems can analyze traffic scenes and provide vital real-time traffic information to drivers and transportation agencies, aiming to enhance road safety and efficiency. These systems also enable authorities to make informed decisions and improve traffic conditions through enhanced situational awareness.

% \begin{figure}[thpb]
%   \centering
%   \includegraphics[width=3.3in]{images/intro.png}
%   \caption{Overview of SKoPe3D Dataset - keypoints, bounding boxes, and different scenes (rain, night, and dawn). }
%   \vspace{-0.2in}
%   \label{fig:intro}
% \end{figure}

\begin{figure}[thpb]
  \centering
\includegraphics[width=3.4in]{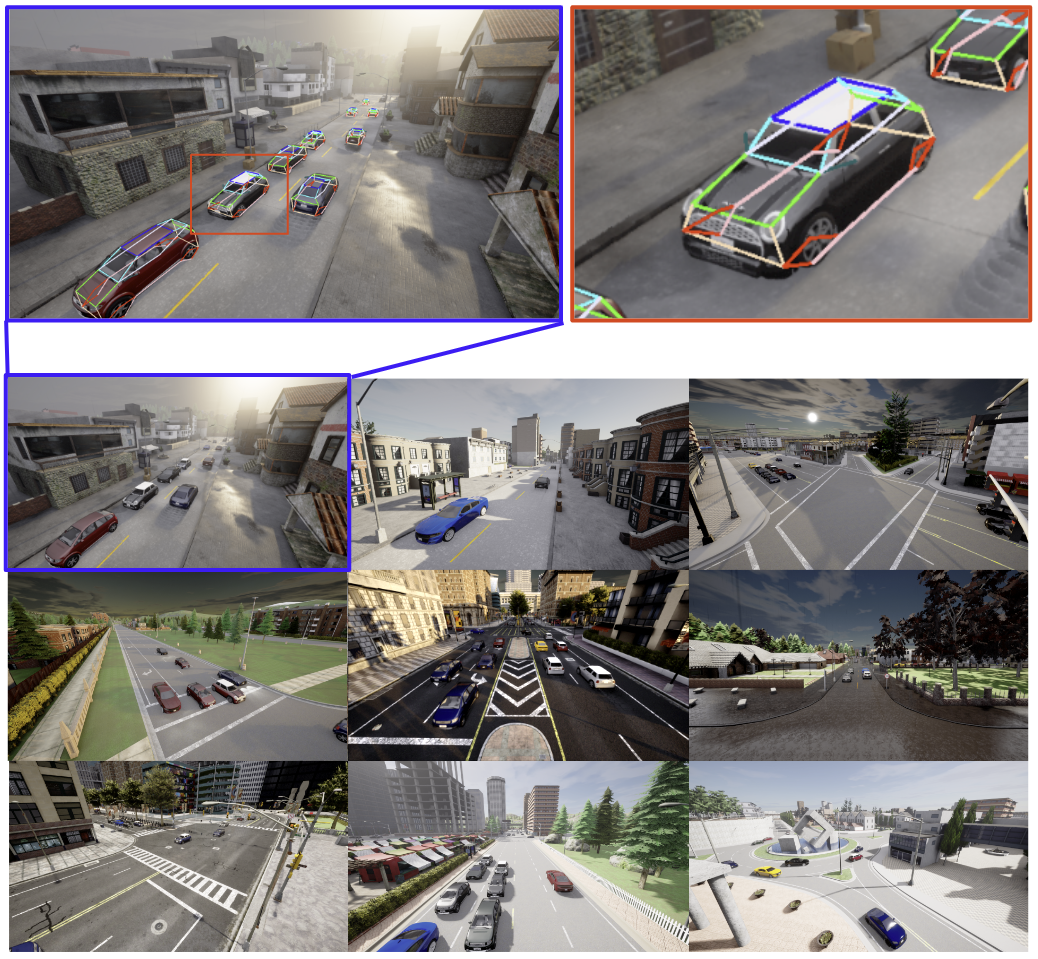}
  \caption{Overview of SKoPe3D Dataset - keypoints, bounding boxes, and different scenes (rain, night, and dawn). }
  \vspace{-0.1in}
  \label{fig:intro}
\end{figure}

 \begin{figure*}[htbp]
  \centering
  \vspace{+0.1in}
  \includegraphics[width=6.8in]{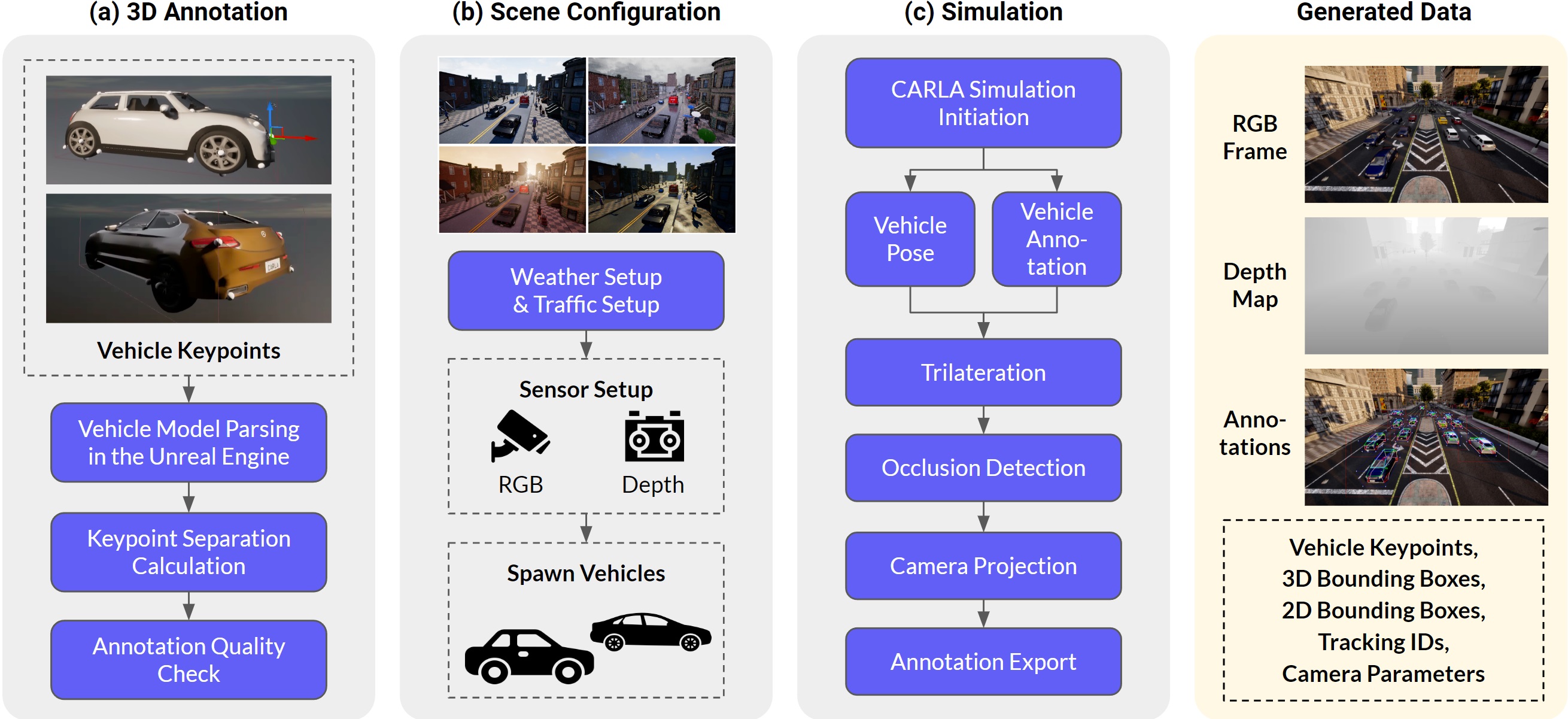}
  \caption{An overview of the SKoPe3D data generation pipeline and the generated data.}
  \vspace{-0.1in}
  \label{fig:arch}
\end{figure*}

Among these vision-based technologies, vehicle detection \cite{sun2006road}\cite{yang2018vehicle}, tracking \cite{choi2015near} \cite{wang2017learning} \cite{xiang2015learning}, and traffic scene reconstruction \cite{zia2013detailed}\cite{lu2021carom}\cite{kar2015category} are heavily researched areas that have seen significant advancement \cite{reddy2018carfusion}. Researchers, regulators, and traffic operators are especially interested in obtaining the vehicle location and orientation from cost-effective monocular cameras. While keypoint-based human pose estimation \cite{newell2016stacked} has emerged as an active research area in recent years resulting in good accuracy, accurate vehicle pose estimation from keypoints remains an outstanding challenge. Specifically, keypoint detection is complicated by factors such as variations in vehicle models and shapes, diversity in camera perspective angles, occlusion, weather, and lighting conditions. This requires datasets that can cover such conditions. However, existing traffic perception models and datasets for vehicle pose estimation predominantly focuses on autonomous driving scenarios, \textit{i.e.}, frontal views captured by ego vehicle-mounted sensors, which limits their applicability to the traffic monitoring cameras on road infrastructure. Hence, for local Departments of Transportation (DoT) and researchers in the field of ITS, it is costly to collect the data, annotate the vehicle keypoints, train keypoint detection models, develop pose estimation algorithms, and deploy such a system with accurate 3D vehicle localization capability.

To address these issues, we present SKoPe3D, i.e., \underline{S}ynthetic dataset and automated data generation pipeline for vehicle \underline{K}eyp\underline{o}int \underline{Pe}ception in the \underline{3D} space. Our dataset is constructed by simulating various scenes in CARLA \cite{dosovitskiy2017carla}. This comprehensive dataset includes generated images with bounding boxes, tracking IDs, and 33 keypoints for each vehicle, as shown in Fig. \ref{fig:intro}. Spanning over {\bf 25K} images across {\bf 28} scenes, SKoPe3D contains over {\bf 150K} vehicle instances and {\bf 4.9 million} keypoints, which is ideal for developing data-driven models for vehicle keypoint detection and pose estimation from a roadside perspective. Besides, our automated data generation pipeline offers a valuable resource for extending our dataset and creating data in new scenes, thereby facilitating the progress of ITS technologies in real-world scenarios. To demonstrate its utility, we trained a keypoint R-CNN model on our dataset as a baseline and conducted a thorough evaluation. Our experiments highlight the dataset's applicability and the potential for knowledge transfer between synthetic and real-world data.

In summary, our contributions are as follows:

\begin{enumerate}

\item We constructed a synthetic dataset with images from roadside camera perspectives at a diverse collection of traffic scenes. Our dataset contains keypoint annotations of each vehicle on every image. The scenes in our dataset were carefully selected to include a range of weather conditions, lighting scenarios, road types, and camera viewpoints.

\item We developed an extension module for the CARLA simulator \cite{dosovitskiy2017carla} and implemented a pipeline for the automated generation of images with 3D annotations of vehicles. This pipeline allows researchers to extend our dataset with more scenes and new vehicles.

\item We validated the performance of Keypoint R-CNN \cite{he2017mask} on our dataset and demonstrated the transferability of knowledge from synthetic data to real-world images.
\end{enumerate}

\section{SKoPe3D Data Generation Pipeline}

Our data generation pipeline is built upon CARLA, which is an open-source simulator designed for research in autonomous driving. However, the simulator does not have built-in support for generating vehicle keypoint annotation. Hence, we developed an extension module based on CARLA Python API to implement our data generation pipeline. The pipeline contains three stages, (a) 3D annotation, (b) scene configuration, and (c) simulation, which are illustrated in Fig. \ref{fig:arch} and detailed in the following three subsections.

\begin{figure*}[]
  \centering
  \vspace{+0.1in}
  \includegraphics[width=6.2in]{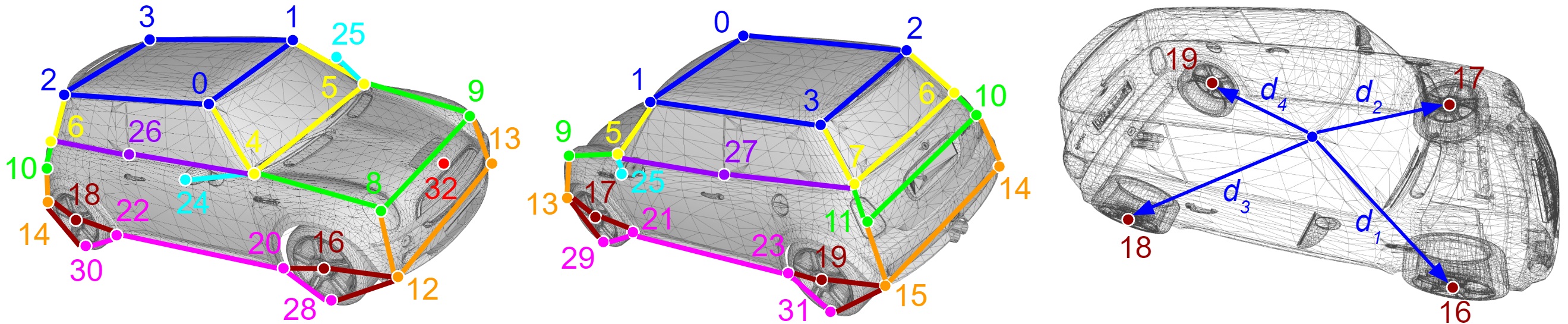}
  \caption{An illustration of vehicle keypoints and trilateration with four reference points.}
  \vspace{-0.2in}
  \label{fig:keypoints}
\end{figure*}

\subsection{3D Keypoint Definition and Annotation}

We defined 33 semantic keypoints for all vehicles, as illustrated in Fig. \ref{fig:keypoints}. This vehicle keypoint scheme generally follows our prior work \cite{lucarom}. These keypoints are designed to capture the most common and strong visual features of different types of vehicles, enabling accurate estimation of the vehicle's pose and shape. All keypoints, except for keypoint 32 (i.e., the keypoint on the manufacturer logo in the front), make full use of geometric relationships and are symmetric in groups of two (left or right) or four (front-right, front-left, rear-right, rear-left), which corresponds to discernible locations of window shield corners, bumper corners, lights, wheels, etc. With these keypoints, a model fitting algorithm can be used to obtain the position, orientation, and shape of a vehicle \cite{lucarom}.

% We defined 33 generic keypoints for all vehicles, as illustrated in Fig. \ref{fig:keypoints}. This vehicle keypoint scheme generally follows our prior work \cite{lucarom}. Although the scheme is initially designed for drone views \cite{lucarom}, we found that they also work for traffic monitoring cameras. All keypoints except the last one are symmetric in groups of two (left or right) or four (front-right, front-left, rear-right, rear-left), which corresponds to import visual features of window shield corners, bumper corners, lights, wheels, etc. These keypoints allow us to obtain accurate information about the positioning and orientation of a vehicle in various view angles. 

To annotate the keypoints in 3D, we edited the vehicle 3D models used by CARLA in Unreal Engine 4, checked their actual dimensions in the real world, and manually placed small white spheres corresponding to our keypoints on the vehicle 3D model, as illustrated in Fig. \ref{fig:arch} (a). The annotation only needs to be done once for each vehicle 3D model. Then, we built a parser to extract the 3D coordinates of our keypoints in the vehicle reference frame. As the keypoints are custom additions, CARLA does not provide direct access to their 3D coordinates in the simulated world reference frame. However, after exploring its API, we found that CARLA can provide the coordinates of the center of mass for each wheel in the world reference frame, which corresponds to our keypoints 16, 17, 18, and 19, as shown in Figure \ref{fig:keypoints}. Hence, we calculated the Euclidean distances between our keypoints and these four reference points and saved them in files so that 3D keypoint coordinates could be recovered during the simulation through trilateration. This step is denoted as keypoint separation in Fig. \ref{fig:arch} (a). Additionally, we developed programs to check the quality of the keypoint annotation on the vehicle 3D model to match its actual dimension.

\subsection{Traffic Scene Configuration}

We manually selected a series of traffic scenes and chose unique viewpoints of cameras on the roadside or the road infrastructure for data generation. In total, we employed seven towns in the simulator and 28 different sites with an RGB camera and a depth camera at each site to simultaneously obtain the image and depth map. Since it is in simulation, the two types of cameras were set up with identical intrinsic parameters and placed at the exact same location and orientation. All cameras had an image resolution of 1920-by-1080 pixels and a field-of-view of 110 degrees. We also consulted experts from the Arizona Institute of Automated Mobility (IAM) \cite{iam} to confirm that our cameras were positioned at realistic places comparable to real-world traffic monitoring cameras deployed in Arizona. The camera parameters were also saved.

Next, we configured each site and each camera for simulation and data recording. The environmental conditions were set by randomizing the weather and time of the day to create a diverse set of video backgrounds, as shown in Fig. \ref{fig:arch} (b). After that, we configured the traffic manager that enables traffic control in the simulator and vehicle spawning with auto-pilot. Vehicles were randomly spawned in each scene with varying color configurations to further improve our dataset's diversity.

\subsection{Simulation and Data Generation}

We developed a Python client program to control the simulation in CARLA frame-by-frame. The images and annotations are automatically generated following a procedure illustrated in Fig. \ref{fig:arch} (c). At each frame, our program queries the simulator's physics engine and obtains the coordinates of the center of mass of the four wheels for all vehicles spawned in the scene. The coordinates of these four reference points are in the world reference frame, which is used to recover the 3D coordinates of vehicle keypoints through a trilateration method, as the example illustrated in Fig. \ref{fig:keypoints} (right). Since a vehicle is a rigid object, this step solves a set of equations \{$(x - x_i)^2 + (y - y_i)^2 + (z - z_i)^2 = d_i^2$\}, where $(x, y, z)$ is the unknown location of a keypoint, $(x_i, y_i, z_i)$ is the known location of a reference point, and $d_i$ is the distance between the keypoint and the reference point obtained from the 3D annotation stage. All 33 keypoints of every vehicle are localized in the world reference frame in this way.

\begin{figure}[thpb]
      \centering
      % \framebox{
      \includegraphics[width=3.0in]{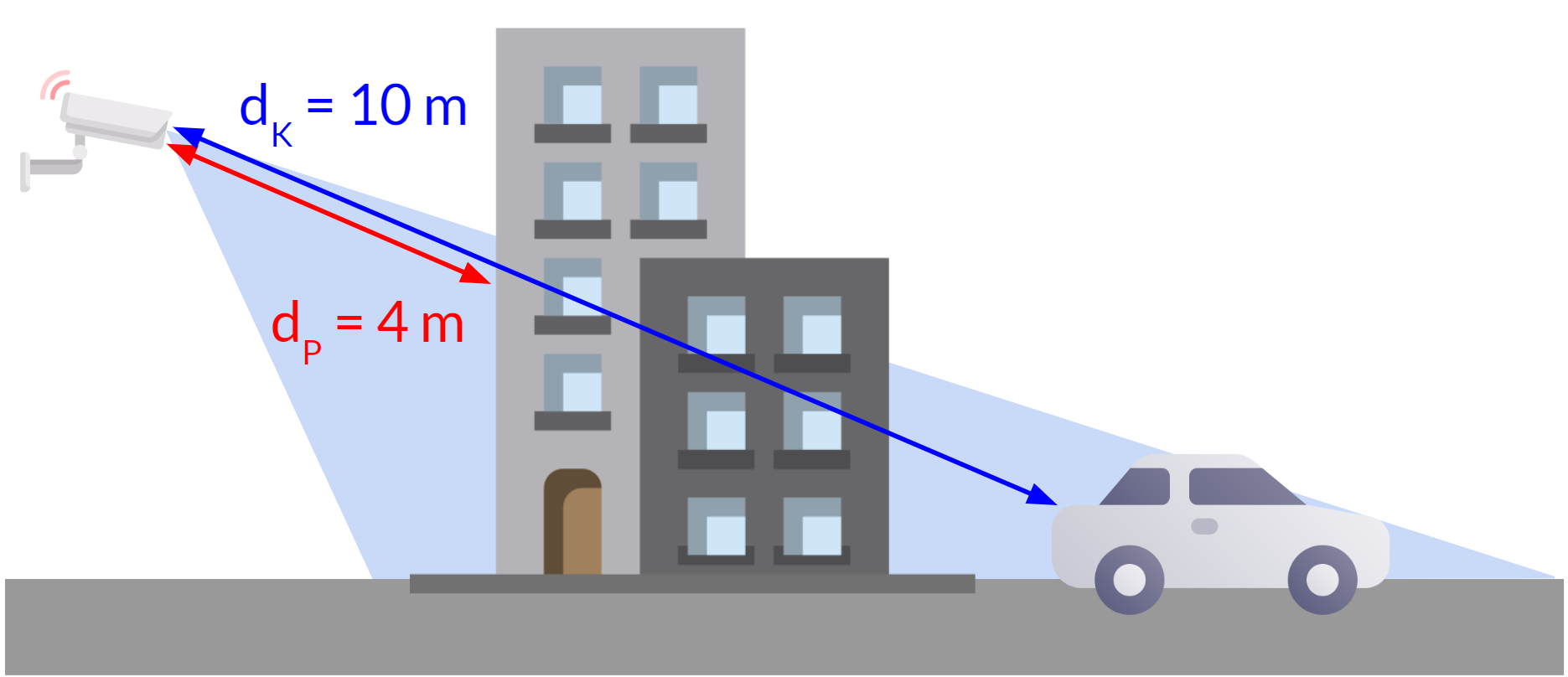}
      % }
      \caption{Illustration of an occlusion scenario.}
      \label{fig:occlusion}
      \vspace{-0.1in}
   \end{figure}

As CARLA does not provide occlusion detection during the simulation, the 3D keypoints are still saved when vehicles are behind the buildings, bridges, or other city infrastructure, as illustrated in Fig. \ref{fig:occlusion}. These vehicles are not visible, and hence, we developed an occlusion detection module using the depth map to remove their keypoints from the generated annotation data, detailed as follows.

\begin{enumerate}
    \item We calculate the Euclidean distance between the camera center and a 3D vehicle keypoint. This distance is denoted as $d_k$.
    
    \item We project the keypoint onto the image reference frame to obtain its pixel coordinates. The actual depth of that pixel is obtained using the depth camera. This distance is denoted as $d_p$.
    
    \item We compare the $d_k - d_p$ against a threshold (20 cm in our implementation) for each keypoint. If more than half of the 33 keypoints exceed the threshold, the target vehicle is considered completely occluded and removed from the annotations.
    
\end{enumerate}

\begin{table*}
\vspace{+0.05in}

\caption{Evaluation results: PCK with $\alpha=0.1$, BB precision-recall and KP precision-recall with varying IoU threshold. }
\vspace{-0.1in}
\begin{center}
\resizebox{\textwidth}{!}{
\begin{tabular}{|c | c | c| c | c | c| c | c| c | c | c| c | c| c | c | c|  } 
 \hline
 \textbf{Scene Split} &  \multicolumn{5}{c|}{\textbf{Easy scene}} &  \multicolumn{5}{c|}{\textbf{Medium scene}} &  \multicolumn{5}{c|}{\textbf{Hard scene}}  \\ 
 \hline

 \multirow{2}{*}{\textbf{\begin{tabular}[c]{@{}c@{}}IoU\\ Threshold\end{tabular}}} &  \multirow{2}{*}{\textbf{\begin{tabular}[c]{@{}c@{}}PCK\\ \%\end{tabular}}}  & \multicolumn{2}{c|}{\textbf{Bounding box}} & \multicolumn{2}{c|}{\textbf{Keypoint}} &  \multirow{2}{*}{\textbf{\begin{tabular}[c]{@{}c@{}}PCK\\ \%\end{tabular}}}  & \multicolumn{2}{c|}{\textbf{Bounding box}} & \multicolumn{2}{c|}{\textbf{Keypoint}} &  \multirow{2}{*}{\textbf{\begin{tabular}[c]{@{}c@{}}PCK\\ \%\end{tabular}}}  & \multicolumn{2}{c|}{\textbf{Bounding box}} & \multicolumn{2}{c|}{\textbf{Keypoint}} \\ 
  &  & Pr & Re & Pr & Re &  & Pr & Re & Pr & Re &  & Pr & Re & Pr & Re  \\ 
 \hline
 \hline

  0.9 & 98.65 & 0.408 & 0.345 & 0.462 & 0.931 & 94.56 & 0.635 & 0.537 & 0.433 & 0.894  & 45.18 & 0.123 & 0.19 & 0.146 & 0.553  \\ 
   \hline
   
  0.8 & 95.59 & 0.793 & 0.641 & 0.488 & 1.000 & 93.77 & 0.941 & 0.776 & 0.475 & 0.983 & 63.05 & 0.382 & 0.615 & 0.366 & 0.990  \\ 
   \hline
   
  0.7 & 95.01 & 0.951 & 0.775 & 0.486 & 1.000  & 95.75 & 0.951 & 0.785 & 0.477 & 0.987 & 64.74 & 0.513 & 0.809 & 0.385 & 1.000   \\ 
   \hline
   
  0.6 & 95.51 & 0.952 & 0.776 & 0.488 & 1.000 & 93.67 & 0.962 & 0.795 & 0.479 & 0.992  & 64.58 & 0.568 & 0.882 & 0.386 & 1.000  \\ 
   \hline

 \hline
\end{tabular}
}
\vspace{-0.2in}
\label{table:metricseasy}
\end{center}
\end{table*}

Finally, the keypoints were projected to the image using the camera parameters, and the 2D keypoint coordinates on the image were exported together with the vehicle bounding boxes and tracking IDs as annotations. An overview of the generated data is shown in Fig. \ref{fig:arch} (rightmost column).

\section{Experimental Evaluation}

Our experiments focus on localizing keypoints of vehicles in synthetic and real-world images captured under diverse settings and scenes, encompassing various viewpoints and road configurations. For this purpose, we utilize Keypoint R-CNN with ResNet50-FPN backbone as a baseline \cite{he2017mask} implemented on PyTorch.

For the sake of efficiency of evaluation, we selected a subset of 13.5K images from our dataset and split them into training and testing sets with a ratio of 8:2. This allows us to train a Keypoint R-CNN model in approximately 12 hours for ten epochs on an NVIDIA TITAN Xp GPU. The training set contains 10.5K images from ten scenes, covering a broad variation of weather conditions, lighting, vehicle visibility, occlusion, as well as the distance between the vehicle and the camera. We use an SGD optimizer with an initial learning rate of 0.001, a momentum of 0.9, a step size of 3, and a Gamma value of 0.2. The testing data contains 3K images from three unseen scenes. These three scenes are named ``easy'', ``medium'', and ``hard'' for convenience.

\subsection{Evaluation Metrics}

We adopt the following metrics proposed in \cite{yang2012articulated} for our evaluation. They are generally recognized in studies on human keypoint detection.

\textbf{Percentage of Correct Keypoints (PCK)}, which measures the number of labeled keypoints that are correctly detected and localized. A predicted keypoint is ``correctly detected'' if its distance to the ground-truth keypoint is equal to or less than $\alpha*L$, where $L=max(height, width)$, i.,e.,  the longer edge of the 2D bounding box, and $\alpha$ is a value between 0 and 1. In our implementation, $\alpha = 0.1$ (following \cite{sanchez2020simple}).

\textbf{Precision and recall of 2D bounding boxes}, which measures the proportion of detected bounding boxes that are correct (i.e., precision) and the proportion of ground-truth bounding boxes that are correctly detected (i.e., recall) given an intersection-over-union (IoU) threshold. A vehicle is considered to be ``detected'' if the detector model reports a confidence score above 0.9 on that vehicle object.
    
\textbf{Precision and recall of keypoints}, which measures the proportion of detected keypoints that are correct (i.e., precision) and the proportion of ground-truth keypoints that are correctly detected (i.e., recall) given the threshold $\alpha*L$ (same threshold in the PCK metric). We did not use a keypoint visibility threshold for the Keypint R-CNN model, and hence, in the calculation of keypoint recall, a keypoint is always detected if the corresponding vehicle is detected.

\subsection{Evaluation Results}

\begin{figure}[!h]  
    \centering
    \includegraphics[width=7cm]{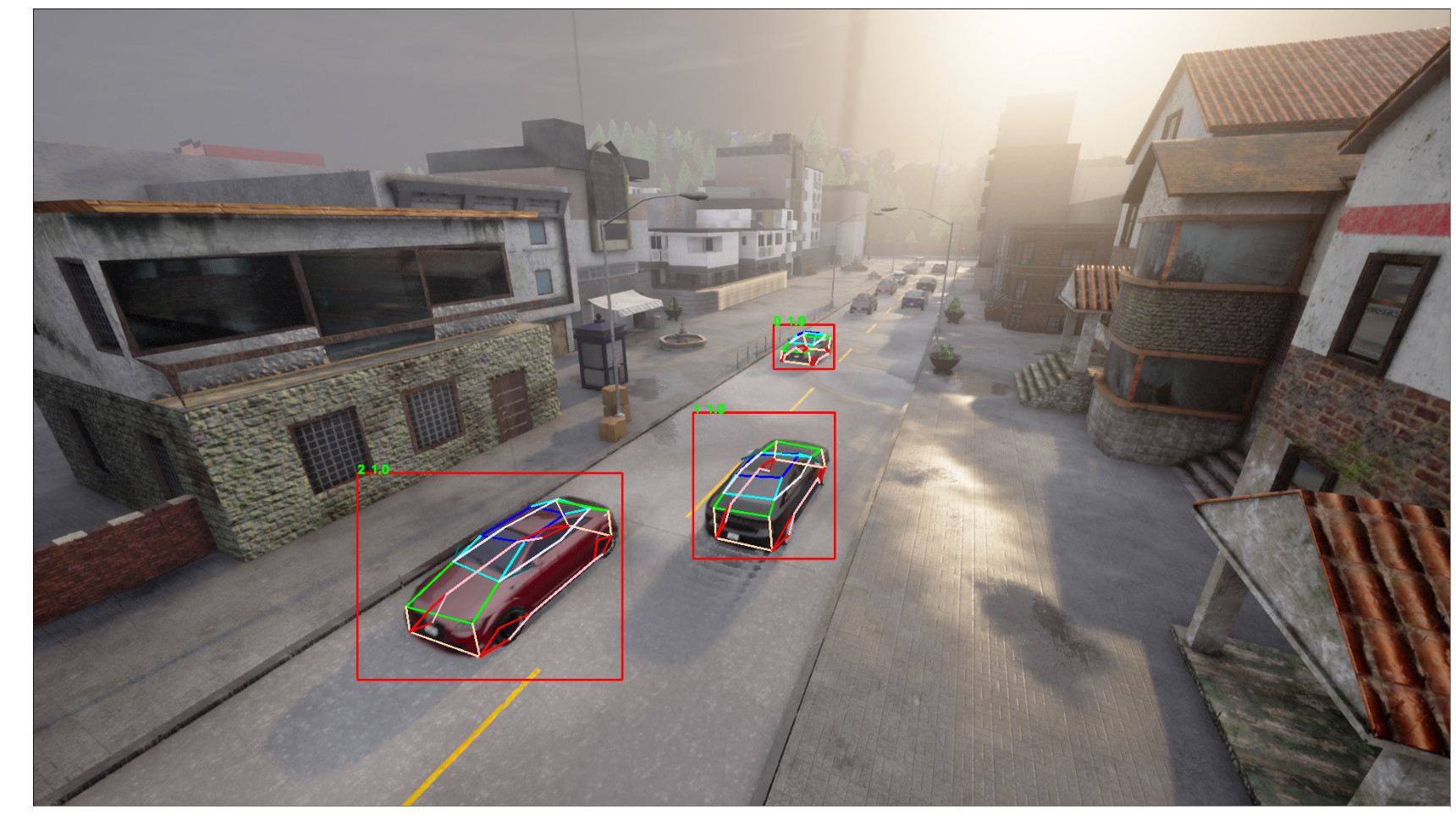}

    \includegraphics[width=4cm]{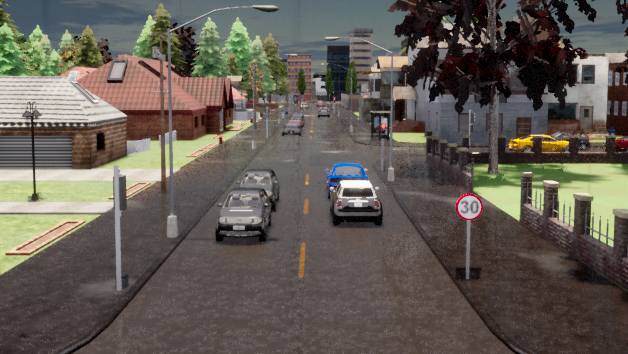}
    \includegraphics[width=4cm]{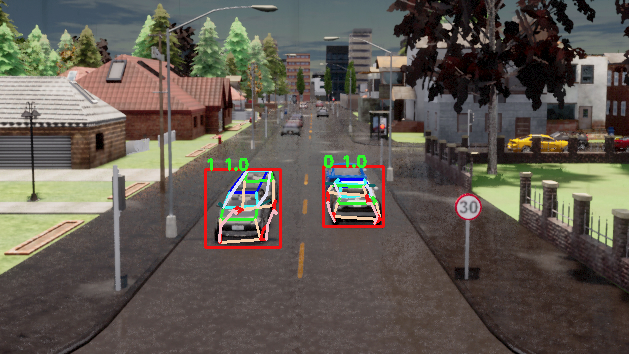}
    \caption{(top) Successful predictions on the easy scene (while vehicles far away are still not detected due to limited image resolution); (bottom) Prediction failures on the hard scene due to vehicle-vehicle occlusion. } 
    \label{fig:good_prediction}
\vspace{-0.1in}
\end{figure}

We trained a Keypoint R-CNN model and ran it on the testing set with the three scenes. TABLE \ref{table:metricseasy} shows the results. The model performed well on the easy and medium scenes and a PCK of 98.65\% and 94.56\% was achieved at 0.9 IoU threshold, while the keypoint precision and recall do not vary much with the IoU threshold. Overall, when vehicles are not partially occluded and within 40 meters from the camera, their keypoints can be reliably detected, as examples shown in Fig. \ref{fig:good_prediction}. However, the hard scene was challenging mainly due to vehicle-vehicle occlusion, low image contrast under unfavorable weather conditions, and lack of resolution when vehicles are far away from the camera, as examples shown in Fig. \ref{fig:good_prediction}. More qualitative results are available in our dataset at \url{https://duolu.github.io/skope3d.html}.

% NOT CROPPED
% \begin{figure}[!h]  
%     \centering
%     \includegraphics[width=7cm]{images/predictedsomething.png}

%     \includegraphics[width=4cm]{images/scene9org.png}
%     \includegraphics[width=4.15cm]{images/scene9predd.png}
%     % \includegraphics[width=10cm]{keypoint loss.png}
%     \caption{(Top) Successful predictions on the easy scene (far away vehicles still are not detected); (Bottom) Prediction failures on the hard scene. } 
%     \label{fig:good_prediction}
% \vspace{-0.1in}
% \end{figure} 

% CROPPED

% \begin{figure}[!h]  
%     \centering
%     \includegraphics[width=7cm]{images/predictedsomething.png}
%     % \includegraphics[width=10cm]{keypoint loss.png}
%     \caption{Successful predictions by the model on the easy scene (far away vehicles still are not detected).} 
%     \label{fig:good_prediction}
% \vspace{-0.1in}
% \end{figure} 

% \begin{figure}[thpb]  
%     \centering
%     % \framebox{
%     \includegraphics[width=4cm]{images/hardsceneorg.png}
%     \includegraphics[width=4.15cm]{images/hardscenepred.png}
%     % }
%     \caption{Prediction failures on the hard scene. } 
%     \label{fig:bad_prediction}
% \vspace{-0.1in}
% \end{figure} 

\begin{table*}
\vspace{+0.05in}
\caption{Comparison to existing datasets. }
\vspace{-0.1in}
\begin{center}
\resizebox{\textwidth}{!}{
\begin{tabular}{|c | c | c | c | c | c | c | c | c| c|c|} 
 \hline
 \textbf{View} & \textbf{Dataset} & \textbf{\begin{tabular}[c]{@{}c@{}}RGB\\ Images\end{tabular}} & \textbf{Scenes} & \textbf{\begin{tabular}[c]{@{}c@{}}LIDAR\\ / Depth\end{tabular}}  & \textbf{\begin{tabular}[c]{@{}c@{}}Key\\ -points\end{tabular}} & \textbf{3D Boxes} & \textbf{2D Boxes} & \textbf{\begin{tabular}[c]{@{}c@{}}RGB Image\\ Resolution\end{tabular}} & \textbf{Year} & \textbf{Diversity}\\ [0.5ex] 
 \hline
\multirow{10}{3em}{Frontal} & KITTI \cite{geiger2012we} & 15K & 22 & Yes & - & 80K & 80K & 1392x512 & 2013 &  \\ 
  % \hline
 & ApolloScape \cite{huang2019apolloscape} & 144K & / & Yes & - & 70K &  & 3384x2710 & 2019 & Night \\ 
 % \hline
 & Lyft Level 5 \cite{lyft2019} & 46K & 366 & Yes & - & 1.3M &  & 1920x1080 & 2019 &  \\ 
 % \hline
 & A2D2 \cite{geyer2020a2d2} & 12K & / & Yes & - & 9K &  & 1928x1208 & 2019 & - \\ 
 & H3D \cite{patil2019h3d} & 27.7K & 160 & Yes & - & 1M &  & 1920x1200 & 2019 & - \\ 
 & Argoverse \cite{chang2019argoverse} & 22K & 113 & Yes & - & 993K &  & 1920x1200 & 2019 & Rain, Night \\ 
 & CityScapes \cite{gahlert2020cityscapes} & 5K & 1150 & Yes & - & 27K &  & 2048x1024 & 2019 & - \\ 
 & nuScenes \cite{caesar2020nuscenes} & 1.4M & 1000 & Yes & - & 1.4M &  & 1600x900 & 2019 & Rain, Night \\ 
 & Waymo Open \cite{sun2020scalability} & 230K & 1150 & Yes & - & 12M & 9.9M & 1920x1080 & 2020 & Rain, Night, Dawn \\ 
 & Apollocar3D \cite{song2019apollocar3d} & 5.5K & 22 & Yes & Yes & 60K &  & 3384x2710 & 2018 & - \\ 
 \hline
\multirow{6}{3.8em}{Roadside} & BoxCars116K \cite{sochor2018boxcars} & 116K & 137 & No & - & 116K &  & 128x128 & 2020 & - \\ 
 & Rope3D \cite{ye2022rope3d} & 50K & 26 & Yes & - & 1.5M &  & 1920x1080 & 2022 & Rain, Night, Dawn \\ 
 & CarFusion \cite{reddy2018carfusion} & 54K & 13 & No & Yes & 100K & 100K & 1920x1080 & 2018 & - \\ 
 & CityFlow-ReID \cite{tang2019cityflow} & 50K & 40 & No & - & 229K &  & 1280x960 & 2019 & - \\ 
 & VeRi-776 \cite{wang2017orientation} & 50K & 20 & No & Yes &  40K &  & 133x152 & 2020 & - \\  
 % & CAROM & 60K & 8 & No & - & - & 186K & 186K & 1280x720 & 2021 & - \\ 
 % \hline
% Drone & CAROM Air & 4.3K & 40 & No & / & 19 & 12K & 12K  &  & 2023 & - \\  
\hline
Roadside & \textbf{SKoPe3D (ours)} & 25K & 28 & Yes & Yes & 151K & 151K  & 1920x1080 & 2023 & Rain, Night, Dawn \\
 \hline
\end{tabular}}
\end{center}
\label{table:datasets}
\vspace{-0.2in}
\end{table*}

\subsection{Generalization to Real-World Scenes}

\begin{figure}[!htb]
    \centering

\includegraphics[width=3.3in]{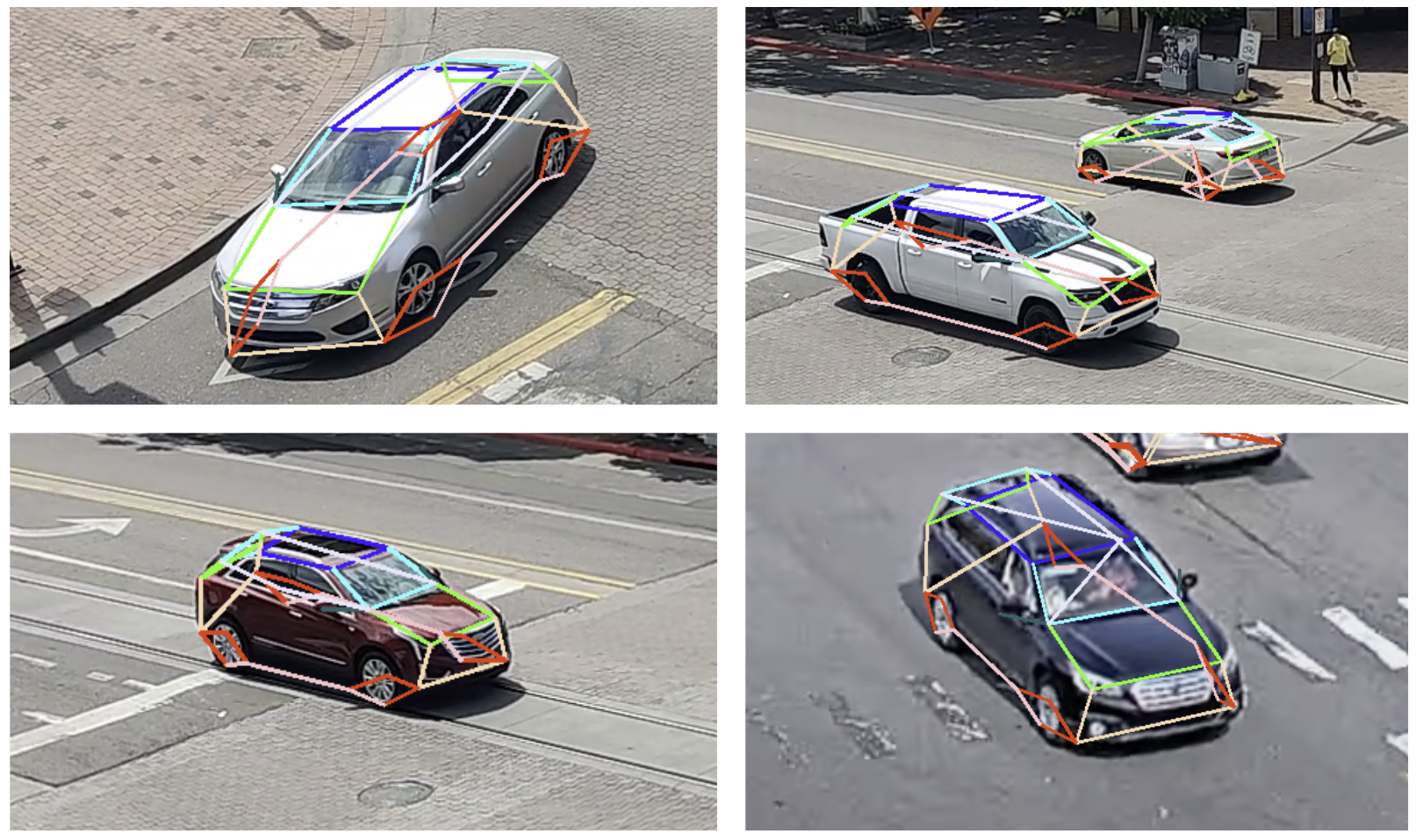}

\caption{Qualitative results on real-world vehicle images.} 
    \label{fig:real_world}
\vspace{-0.1in}
\end{figure}

We qualitatively evaluated our Keypoint R-CNN model on real-world traffic monitoring videos captured at an intersection in Tempe, Arizona, as depicted in Fig. \ref{fig:real_world}. The two images on the right of this figure show the model's capability to detect multiple vehicles and generalize to pickup trucks that are absent in the training data. However, it should be noted that such a keypoint detection model trained on our synthetic dataset has limited detection accuracy in practice due to the sim-to-real gap. Our main goal is to use the synthetic dataset to pre-train a keypoint detector and fine-tune it on a small set of annotated real-world images so as to reduce the cost of developing such a detector by annotating a large-scale dataset of real-world images.

% \begin{figure}[!htb]
%     \centering

% % \framebox{
% \minipage{0.22\textwidth}
% \includegraphics[width=\linewidth]{images/sedan.jpeg}
% % \caption{genworkflow}\label{fig:awesome_image1}
% \endminipage
% \minipage{0.22\textwidth}
% \includegraphics[width=\linewidth]{images/multiple.jpeg}
% % \caption{pyramidprocess}\label{fig:awesome_image2}
% \endminipage\hfill
% \minipage{0.22\textwidth}%
% \includegraphics[width=\linewidth]{images/suv.jpeg}
% % \caption{mt-simtask}\label{fig:awesome_image3}
% \endminipage
% \minipage{0.22\textwidth}
% \includegraphics[width=\linewidth]{images/truck.jpeg}
% % \caption{pyramidprocess}\label{fig:awesome_image2}
% \endminipage
% % }
% % \caption{Sample Frames Were Extracted from a Smartphone Video of the Mill Avenue Intersection, Tempe, Az. } 
% \caption{Qualitative results on real-world traffic images.} 
%     \label{fig:real_world}
% \vspace{-0.1in}
% \end{figure}

% \begin{figure}[!htb]
%     % \centering

% % \framebox{
% \minipage{0.22\textwidth}
% \includegraphics[width=\linewidth]{images/1.jpeg}
% % \caption{genworkflow}\label{fig:awesome_image1}
% \endminipage
% \minipage{0.22\textwidth}
% \includegraphics[width=\linewidth]{images/2.jpeg}
% % \caption{pyramidprocess}\label{fig:awesome_image2}
% \endminipage\hfill
% \minipage{0.22\textwidth}%
% \includegraphics[width=\linewidth]{images/3.jpeg}
% % \caption{mt-simtask}\label{fig:awesome_image3}
% \endminipage
% \minipage{0.22\textwidth}
% \includegraphics[width=\linewidth]{images/4.jpeg}
% % \caption{pyramidprocess}\label{fig:awesome_image2}
% \endminipage
% % }
% % \caption{Sample Frames Were Extracted from a Smartphone Video of the Mill Avenue Intersection, Tempe, Az. } 
% \caption{Examples of qualitative results on real-world traffic monitoring images.} 
%     \label{fig:real_world}
% \vspace{-0.1in}
% \end{figure}

\section{Comparison to Existing Datasets}

Due to active research in autonomous driving and traffic monitoring, there’s an increase in large-scale traffic scene datasets as listed in TABLE \ref{table:datasets}. The datasets are divided based on the camera view, \textit{i.e.}, ``frontal'' from an ego vehicle or ``roadside'' from a traffic monitoring camera. In the table, '/' denotes an unknown value, while '-' denotes that the information is unavailable.

Frontal-view datasets are primarily used in autonomous driving use cases. The KITTI Vision Benchmark Suite, released in 2012, was one of the first datasets for self-driving use cases. It provided multimodal data and opened up various challenges in autonomous driving. Other large-scale datasets took inspiration from KITTI and improved the quality of autonomous driving datasets like H3D, ApolloScape and ApolloCar3D, Waymo Open Dataset, Argoverse, Lyft Level 5, A2D2, and many more. These datasets led to advanced research in autonomous driving, but they were from an ego-vehicle perspective.

Roadside perspective datasets, on the other hand, can help with 3D localization from a roadside surveillance camera perspective \cite{altekar2021infrastructure}. However, the availability of such datasets is limited. A few datasets in this category are CityFlow  \cite{tang2019cityflow}, CAROM \cite{lu2021carom}, BoxCars \cite{sochor2018boxcars}, Rope3D \cite{ye2022rope3d}, and PASCAL3D+ \cite{xiang2014beyond} dataset. Most of these datasets lack keypoint annotations for the vehicles in the scene. The ApolloCar3D dataset has keypoint annotations, but it's from a frontal-view perspective. The CarFusion \cite{reddy2018carfusion} dataset has keypoint annotations, but it was recorded by people holding their smartphones while walking on the sidewalk. Therefore, it cannot be considered a traffic monitoring dataset. The veri-776 dataset \cite{wang2017orientation}\cite{gupta2021vehipose} has cropped image patches of vehicles from traffic monitoring videos with keypoint annotations. It doesn't have annotations on the entire scene, similar to ApolloCar3D. 

Our dataset is synthetic with vehicle keypoints, which makes SKoPe3D distinct from existing open traffic monitoring datasets. There are several open-source simulation environments, such as CARLA \cite{dosovitskiy2017carla}, LGSVL \cite{rong2020lgsvl}, AirSim \cite{shah2018airsim}, WorldGen \cite{deep2022worldgen}, etc. CARLA Simulator stands out among them due to its advanced state-of-the-art rendering quality, realistic physics, sensor library and capability to generate believable behaviors for vehicles and pedestrians. This makes it a suitable platform for SKoPe3D. There are also similar synthetic datasets \cite{herzog2023synthehicle}\cite{hu2022sim}, but they do not have vehicle keypoints. Additionally, our data generation software module will be made openly available to allow interested researchers to extend our dataset or generate their own data. Furthermore, since our data is simulated without real-world sensitive information such as license plates or human faces, SKoPe3D does not have privacy issues, making it easily available for the traffic research community.

\section{Conclusion and Future Work}

This paper presents SKoPe3D, a synthetic dataset and data generation pipeline for vehicle keypoint perception in 3D, which can benefit various roadside perception tasks, including vehicle pose estimation, vehicle tracking, and 3D traffic scene reconstruction. We also conducted a baseline evaluation of the dataset, which demonstrated the dataset's usability. Furthermore, this study represents a step towards bridging the gap between synthetic and real-world data in vehicle keypoint detection.

% This study also represents a significant step towards bridging the gap between synthetic and real-world data. Training a keypoint detector on a synthetic dataset showcased the model's ability to generalize to some extent on real-world images, which highlights the knowledge transferability between synthetic and real-world data, underscoring the value of SKoPe3D as a crucial resource for advancing research and applications in intelligent transportation systems. The availability of SKoPe3D dataset and the insights gained from our experiments contribute to the ongoing efforts in developing robust and accurate vehicle keypoint detection algorithms, thereby facilitating advancements in various transportation-related domains.
In future work, we plan to investigate accurately estimating vehicles' shape, position, and orientation using detected keypoints and model fitting. We also aim to conduct a thorough quantitative analysis of real intersection videos by manually annotating them and evaluating the keypoint detector trained on SKoPe3D dataset. The availability of SKoPe3D dataset and the insights gained from our experiments contribute to the ongoing efforts in developing robust and accurate vehicle keypoint detection algorithms, thereby facilitating advancements in various transportation-related domains.


%%%%%%%%%%%%%%%%%%%%%%%%%%%%%%%%%%%%%%%%%%%%%%%%%%%%%%%%%%%%%%%%%%%%%%%%%%%%%%%%

%%%%%%%%%%%%%%%%%%%%%%%%%%%%%%%%%%%%%%%%%%%%%%%%%%%%%%%%%%%%%%%%%%%%%%%%%%%%%%%%

%%%%%%%%%%%%%%%%%%%%%%%%%%%%%%%%%%%%%%%%%%%%%%%%%%%%%%%%%%%%%%%%%%%%%%%%%%%%%%%%
%\section*{APPENDIX}


%\section*{ACKNOWLEDGMENT}

% The preferred spelling of the word ÒacknowledgmentÓ in America is without an ÒeÓ after the ÒgÓ. Avoid the stilted expression, ÒOne of us (R. B. G.) thanks . . .Ó  Instead, try ÒR. B. G. thanksÓ. Put sponsor acknowledgments in the unnumbered footnote on the first page.

%%%%%%%%%%%%%%%%%%%%%%%%%%%%%%%%%%%%%%%%%%%%%%%%%%%%%%%%%%%%%%%%%%%%%%%%%%%%%%%%

% References are important to the reader; therefore, each citation must be complete and correct. If at all possible, references should be commonly available publications. 

% \begin{thebibliography}{99}

% \end{thebibliography}

\bibliographystyle{IEEEtran}
\bibliography{dis}
% \bibliography{dis}
\end{document}